\documentclass[runningheads]{llncs}
\usepackage[colorlinks=true,linkcolor=blue,citecolor=blue,urlcolor=blue]{hyperref}
\usepackage[T1]{fontenc}
\usepackage{graphicx}
\usepackage{tcolorbox}
\usepackage{amsmath}
\usepackage{amssymb}
\usepackage{multirow}

\newcommand{\etal}{\textit{et al. }} 
\begin{document}
\title{LLMs for Cardiovascular Risk Prediction from Structured Clinical Data}
%
%
\author{Jeba Maliha \and
Md Rafiul Kabir
}

%
\institute{Central Michigan University, Mount Pleasant, MI, USA \\
\email{malih1j@cmich.edu, kabir2m@cmich.edu}\\
}
\maketitle              
\begin{abstract}
Coronary artery disease (CAD) remains one of the leading causes of death globally, highlighting the need for reliable predictive systems to support early diagnosis and risk assessment. While traditional machine learning models perform well on structured clinical data, large language models (LLMs) present new possibilities to interpret medical information expressed in natural language. In this work, we develop a hybrid framework that bridges structured clinical data and natural-language representations for CAD prediction. Using a publicly available dataset of 1,190 patient records with 11 clinical attributes, structured variables are converted into interpretable feature representations and synthetic clinical narratives using LLMs. A validation pipeline performs reverse extraction of clinical variables and computes a consistency score with the original records, achieving an average fidelity of 94.61\%. We then evaluate four conventional machine learning models and compare their performance with LLM-based classification under zero-shot and few-shot prompting settings. We use two LLMs here, GPT and Gemini. Experimental results show that Random Forest achieves the highest accuracy. Despite this advantage, LLM-based classification remains beneficial in real-world clinical settings. This is because LLMs operate directly on natural language patient descriptions, meaning that sensitive
numerical patient data such as exact lab values, blood pressure readings, and diagnostic codes are kept private. Findings suggest that combining structured clinical data with LLM-generated narratives can enable new directions for hybrid clinical prediction systems.

\end{abstract}
\section{Introduction}

Coronary artery disease (CAD) is one of the leading causes of mortality worldwide and represents a major public health challenge \cite{shao2020coronary}. Early detection and accurate risk prediction are critical for improving patient outcomes and enabling timely clinical intervention. Over the past decade, machine learning techniques have been increasingly applied to clinical datasets to support automated cardiovascular risk prediction. Traditional models such as Logistic Regression, Support Vector Machines, Random Forest, and Gradient Boosting have demonstrated strong performance when trained on structured clinical variables, including age, blood pressure, cholesterol levels, and electrocardiographic measurements \cite{baghdadi2023advanced,gururaj2025recent}.
Despite these advances, most existing approaches rely exclusively on structured numerical data. However, in real clinical settings, patient information is often documented in natural language through electronic health records (EHRs), physician notes, and diagnostic reports. This mismatch between structured machine learning inputs and narrative clinical documentation limits the applicability of many predictive models. Recent developments in large language models (LLMs) \cite{mehdi2025llm} offer new opportunities to bridge this gap by enabling automated generation and interpretation of natural language medical descriptions.

\begin{figure}
    \centering
    \includegraphics[width=0.98\linewidth]{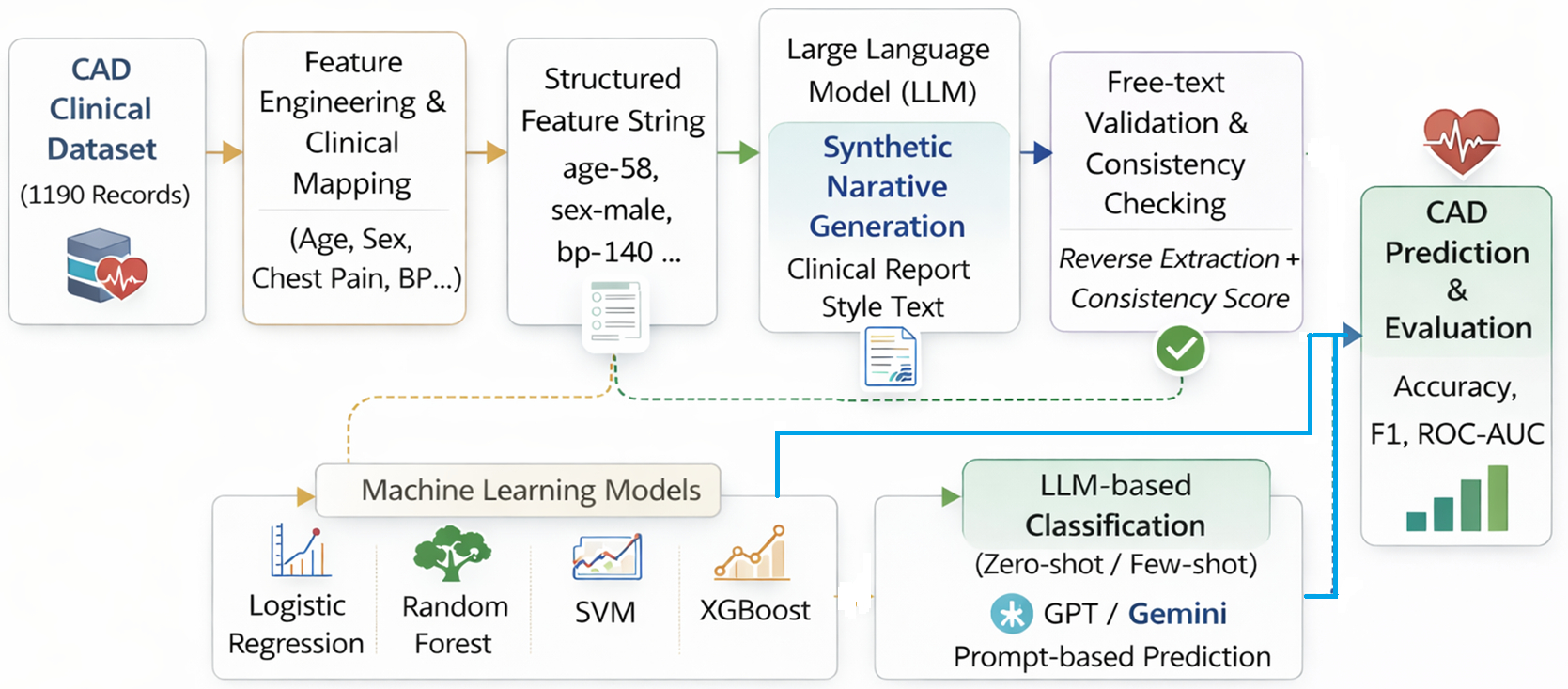}
    \caption{Workflow for coronary artery disease prediction using ML and LLM}
    \label{fig:8}
\end{figure}
In this work, we propose a hybrid framework (shown in Fig. \ref{fig:8}) that combines structured clinical data, synthetic clinical narrative generation, and machine learning for coronary artery disease prediction. Starting from a publicly available CAD dataset containing 1,190 patient records and 11 clinical attributes, structured variables are first transformed into interpretable feature representations. These features are then used to generate synthetic clinical narratives using a large language model. To ensure factual accuracy, we introduce a validation pipeline that performs reverse extraction of clinical values from the generated narratives and computes a consistency score with the original structured records.
Using the validated dataset, we evaluate four conventional machine learning models --- Logistic Regression, Random Forest, Support Vector Machine, and XGBoost --- and compare their performance with LLM-based classification under zero-shot and few-shot prompting conditions. Our results show that ensemble-based models achieve the highest predictive performance, while LLM-based approaches demonstrate promising capability in interpreting natural-language clinical descriptions.

The main contributions of this work are summarized as follows:

\begin{itemize}
    \item We propose a hybrid framework that integrates structured clinical data, large language models, and machine learning for coronary artery disease prediction.
    \item We introduce a synthetic clinical narrative generation pipeline that converts structured patient records into natural-language clinical reports.
    \item We develop a validation mechanism that ensures factual consistency between generated narratives and original clinical data.
    \item We provide a comparative evaluation of traditional machine learning models and LLM-based classification approaches on the structured clinical data and validated dataset.
\end{itemize}

\section{Related Work}
Recent studies have explored the integration of natural language processing, LLMs, and machine learning techniques for cardiovascular disease prediction and clinical data analysis.
Yang et al. \cite{yang2025llm} proposed an LLM-augmented NLP pipeline for cardiovascular disease prediction using unstructured patient narratives. Their approach uses BioClinicalBERT to generate contextual embeddings from symptom descriptions, which are then classified using a Random Forest model, achieving 85.7\% accuracy and an F1-score of 85.3\% on a synthetic dataset without domain-specific fine-tuning. Li et al. \cite{li2025ai} proposed a hybrid AI framework that combines NLP and machine learning for cardiovascular risk prediction in athletes. Their method uses a RoBERTa-based model to extract features from unstructured clinical narratives, which are then processed by a Wolf Pack Search–optimized Dynamic Random Forest (WPSA-DRF). The approach achieved 92.5\% accuracy, 99.23\% recall, and 98.1\% AUC-ROC, outperforming several baseline models. 
Chen \etal \cite{chen2025large} proposed a lightweight dual-attention ECG network for heart failure (HF) risk prediction using 12-lead electrocardiograms. Their model incorporates cross-lead and lead-specific temporal attention modules and employs LLM-based pretraining for ECG report alignment, achieving improved predictive performance on U.K. Biobank cohorts, with C-index scores of 0.6349 (UKB-HYP) and 0.5805 (UKB-MI).

Another recent work by Wang et al. \cite{wang2025boosting} proposed a hybrid AI model that integrates a large language model with a traditional machine learning classifier for heart disease prediction. In their approach, the LLM generates a natural-language summary and a risk score from raw clinical data, which is then combined with the original numerical features. This augmented feature set is subsequently used by an XGBoost classifier to produce the final prediction. 
Pan \etal \cite{pan2025integrating} developed an LLM–based pipeline to identify multiple diseases from electronic health record clinical notes without requiring extensive manual labeling. When applied to a cohort of 3,088 patients with more than 551,000 notes, the approach demonstrated improved sensitivity compared with ICD code–based methods for detecting conditions such as acute myocardial infarction, diabetes, and hypertension. Mila \etal \cite{Mila2026ScreeningPaL} presented a text-based method for early autism risk detection using caregiver-reported descriptions, training language models on synthetic free-text and achieving up to 90\% accuracy with fine-tuned transformers outperforming other approaches. They further show that augmenting data with realistic noise improves recall and generalization, enabling a low-cost, accessible screening approach that can support early intervention and specialist evaluation.

\section{Methodology}

\subsection{\textbf{Dataset Construction and Feature Engineering}}

We used a publicly available coronary artery disease (CAD) dataset consisting of tabular clinical data commonly used in cardiovascular risk assessment. The dataset includes physiological attributes such as age, sex, chest pain type, resting blood pressure, cholesterol levels, fasting blood sugar, resting electrocardiogram (ECG) results, maximum heart rate achieved, exercise-induced angina, ST depression (oldpeak), and ST slope \cite{mondal2026ai}. The target variable represents the presence or absence of heart disease. These variables correspond to established clinical indicators frequently employed in CAD diagnosis. The goal is to obtain paired tabular and textual representations of each patient that are suitable for both conventional machine‑learning models and language-model-based approaches.
Let the original structured dataset be:



\[
\mathcal{D} = \{(x_i, y_i)\}_{i=1}^{N}, 
\quad x_i \in \mathcal{X} \subseteq \mathbb{R}^p,
\quad y_i \in \{0,1\}
\]

To make the dataset more interpretable for both clinicians and language models, we convert each coded variable into a human‑readable clinical category. For example, the binary sex indicator is mapped to “female” or “male”, chest pain types are mapped to “typical angina”, “atypical angina”, “non‑anginal pain”, or “asymptomatic”, and ECG findings are mapped to “normal ECG”, “ST–T wave abnormality”, or “left ventricular hypertrophy”. Similar mappings are defined for fasting blood sugar status, presence of exercise‑induced angina, and the qualitative slope of the ST segment. These mapping tables are stored to create descriptive strings.

\subsection{Synthetic Clinical Narrative Generation}
\subsubsection{Structured Text Creation:}
For every row, the previously constructed “Feature” string is parsed to retrieve the mapped clinical values, and these values are injected into a carefully designed prompt. For each patient record, LLM is first prompted with the raw numerical values to produce a single-line, comma-separated structured summary in a consistent feature-value pair format (e.g., "age-40, sex-male, chest pain type-non-anginal pain, resting bp-140 mm Hg, cholesterol-289 mg/dL..."), with temperature set to 0 for fully deterministic output. To handle any API errors, a backup function is built into the pipeline that performs the same clinical mapping locally, so the process keeps running without interruption. Once the feature string is ready, it is passed into a second prompt. 

\begin{figure}
    \centering
    \includegraphics[width=0.99\linewidth]{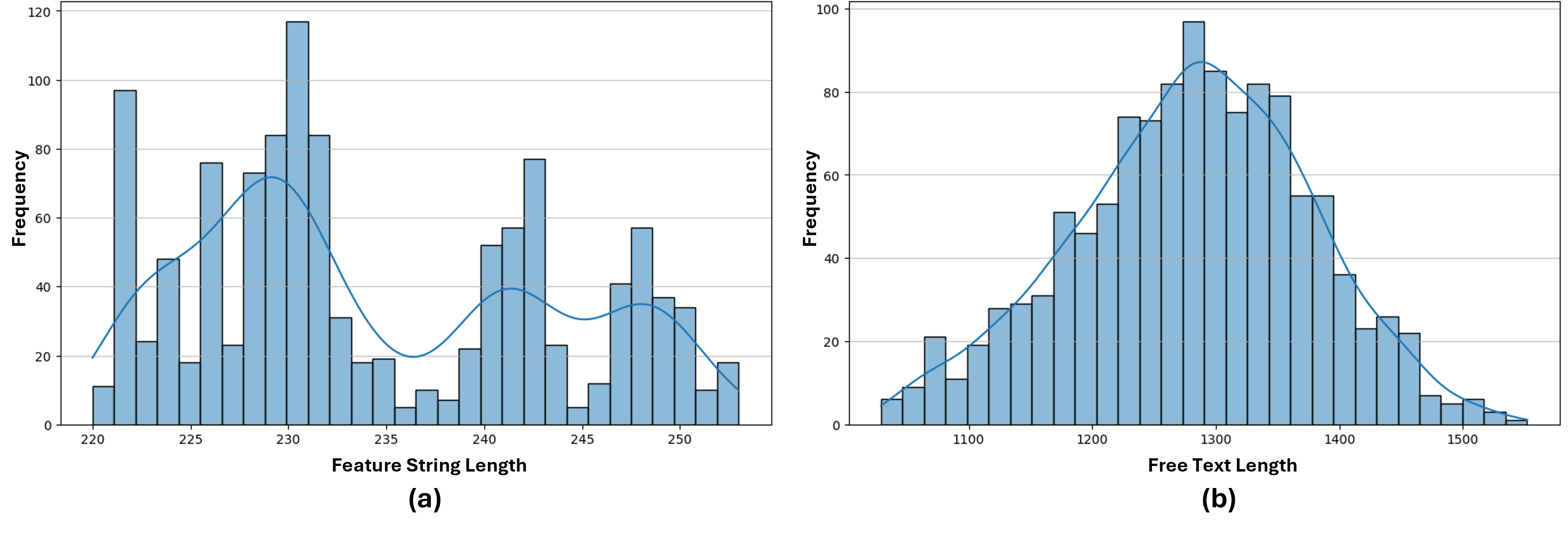}
    \caption{Distribution of (a) Feature String Lengths and (b) Free Text Lengths}
    \label{fig:2}
\end{figure}

\subsubsection{Structured Data to Free-Text Transformations:}
We transformed structured numerical patient records from the heart disease dataset into two distinct forms of readable clinical text shown in Fig \ref{fig:2}. For each patient record, the system produces two separate outputs. The first is a structured summary that lists all clinical feature strings in a consistent format. The second is a detailed paragraph written in the style of a clinical report that describes the same patient in full sentences like how a doctor would write a note in an Electronic Health Record (EHR). The model is instructed to naturally include all clinical values in the report without inventing any extra measurements that are not in the original data. To allow some variation in writing style while keeping the medical content accurate, the temperature is set to 0.3.  Both text formats are generated using an LLM rather than simple rule-based templates. For each patient, both texts are saved together with a label showing whether heart disease is present or not. Any report that is failed to generate or came back empty is marked separately for re-generation from the dataset. The reason for creating two formats is straightforward. The structured summary works well for models that prefer consistent input patterns, while the free-text narrative suits models that need natural medical language. Having both formats makes the dataset flexible enough to support a wider range of natural language processing and evaluation scenarios.

\subsection{Narrative Validation and Quality Assurance}\subsubsection{Free-Text Validation and Correction:}
Each generated free-text narrative is compared against its corresponding original patient record by prompting the LLM to act as a medical data validator. The model identifies any missing values or factual contradictions in the narrative, and if any issues are found, then it produces a corrected version that accurately reflects all original clinical values, preserving natural language flow. The result is returned as a flag indicating whether issues were detected or not.

\subsubsection{Reverse Mapping and Clinical Value Extraction:}
To objectively measure how faithfully the corrected narrative encodes the original data, the LLM is prompted a second time to extract all clinical values from the narrative text and map them back to their original numerical codes. This reverse extraction produces a structured record mirroring the original dataset schema, which enables a direct field-by-field comparison between the source record and the text-derived values.

\subsubsection{Consistency Scoring:} 
To measure how well each narrative matches the original data, we directly compare every extracted feature with its ground‑truth value. Categorical fields have to agree exactly, while numeric fields are allowed a small margin of error (we treat them as matching if they differ by less than 0.5, to absorb minor rounding in the text). For each patient, we then calculate a consistency score by dividing the number of matching fields by the total number of fields and multiplying by 100, shown in Equation \ref{eq:consistency}. This gives a simple percentage that tells us how faithful each narrative is to the underlying clinical record.
For each record, a consistency score is computed as:

\begin{equation}
\text{Consistency\;Score} =
\left(
\frac{\text{Number\;of\;Matched\;Fields}}
{\text{Total\;Number\;of\;Fields}}
\right) \times 100
\label{eq:consistency}
\end{equation}
This produces a percentage-based measure for every sample.

\begin{figure}
    \centering
    \includegraphics[width=0.75\linewidth]{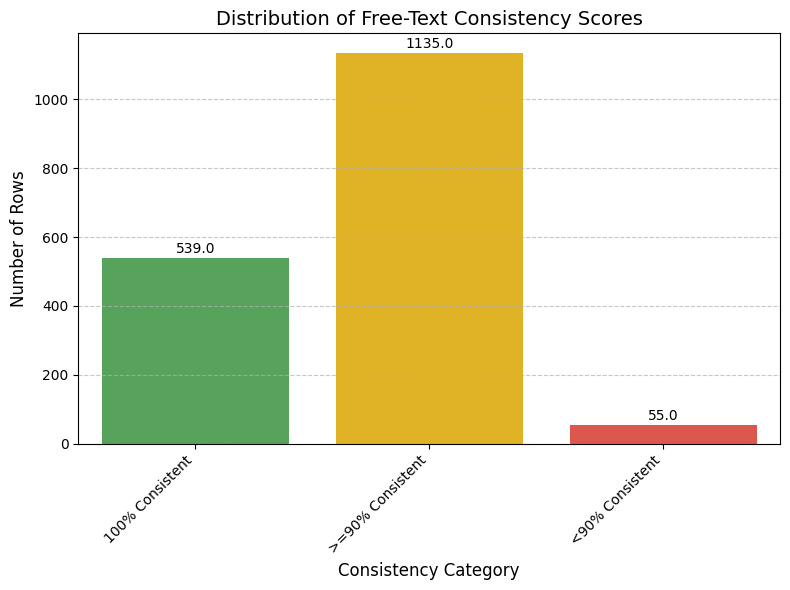}
    \caption{Free-text Consistency Score}
    \label{fig:fig2}
\end{figure}

\subsubsection{Validation Results and Summary Statistics:}
Across all 1,190 patient records, our pipeline achieved an average consistency score of 94.61\%. Of these, 539 records (45.3\%) achieved a perfect score of 100\%, and 1,135 records (95.4\%) scored at or above 90\%, as shown in Fig \ref{fig:fig2}. Only 55 records (4.6\%) fell below the 90\% threshold, indicating a high degree of factual fidelity between the generated narratives and the original structured clinical data.

\subsection{Predictive Modeling}
\subsubsection{Machine Learning Models for CAD Prediction from Clinical Features:}
We use a clinical dataset of 1,190 patient records to train and evaluate machine learning models for predicting coronary artery disease, with 11 clinical features. Before training, we explore the data through correlation analysis and distribution plots to understand how clinical measurements relate to disease outcomes and confirm that no missing values are present and the class distribution is reasonably balanced. The data is then split into training and testing sets using an 80–20 stratified ratio, and all features are standardized to ensure each variable contributes equally during model training. We train and compare four well-known classification algorithms: Logistic Regression, Random Forest, Support Vector Machine, and XGBoost. Every model is evaluated consistently. The evaluation and parameter optimization of these models are described in detail later in Section 4. Machine Learning Models, and their comparative performance, are presented in the Results and Discussion section.

\subsubsection{LLM-based CAD Prediction from free-text:}

We explore how well LLMs can identify coronary artery disease from natural language descriptions of patient clinical data\cite{ahmed2025primer}, without any traditional model training. The dataset is represented as free-text patient summaries paired with binary disease labels. All labels are standardized to a consistent format, and the data is split into training and testing records to evaluate uniformly across all experiments. We design four prompting conditions to systematically assess the effect of in-context learning: a zero-shot setting where the model receives no prior examples and must rely entirely on its own medical knowledge, and three few-shot settings where 5, 10, and 15 balanced labeled patient examples are provided. In all conditions, the model is instructed to respond strictly with one of the two predefined class labels, and the temperature is set to 0 to ensure consistent and deterministic outputs across all runs.

\section{Modeling framework for Coronary Artery Disease Prediction}

\subsection{Machine Learning Models}

\begin{figure}
    \centering
    \includegraphics[width=0.999\linewidth]{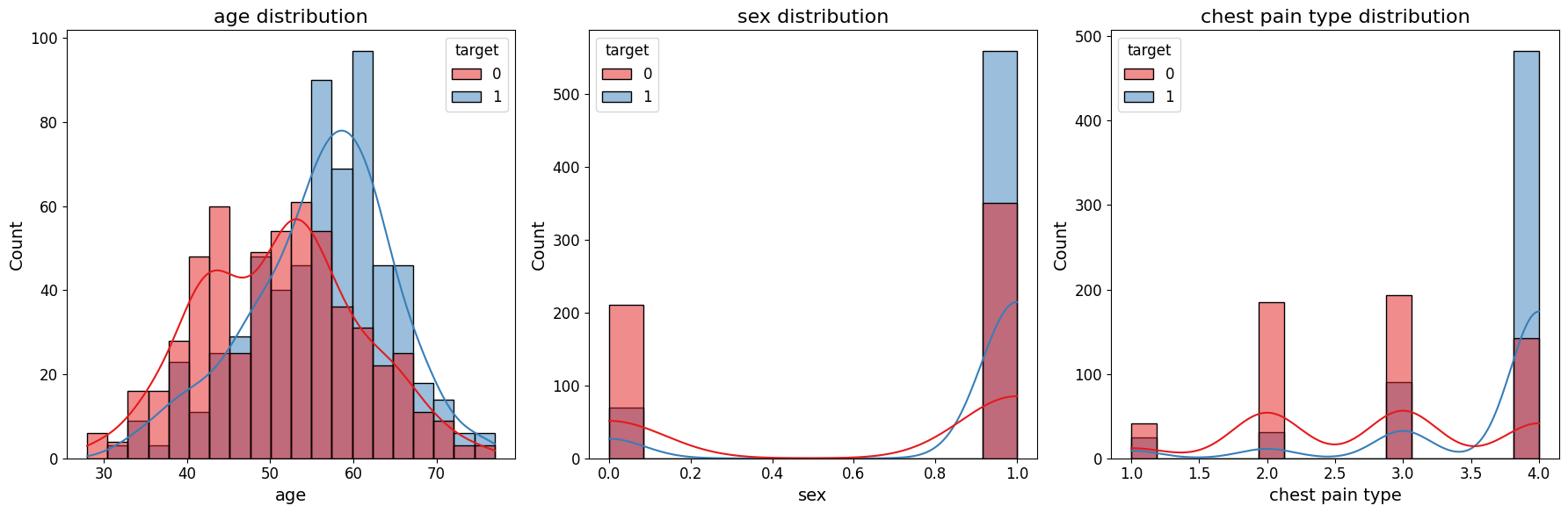}
    \caption{Selected Feature Distributions}
    \label{fig:fig6}
\end{figure}

We design an experimental setup that covers everything from initial data exploration to final model evaluation. The dataset used in our experiments consists of 1,190 patient records, each described by 11 clinically meaningful attributes such as age, chest pain type, resting blood pressure, cholesterol levels, ST slope, and exercise-induced angina, along with a binary label indicating whether the patient has CAD or not. Before any model is trained, we carefully explore the data by computing descriptive statistics \cite{dinh2019data}, examining class balance, and visualizing feature distributions, shown in Fig \ref{fig:fig6} and correlations to better understand the underlying patterns. The data is then preprocessed by splitting it into training and testing sets using an 80–20 stratified split and applying standard scaling to ensure all features contribute equally during model training.
We then train and compare four well-established supervised learning algorithms, namely Logistic Regression, Random Forest, Support Vector Machine, and XGBoost. Each of them is chosen for its unique strengths in handling clinical classification tasks. Every model is evaluated consistently using accuracy, precision, recall, F1-score, ROC-AUC, and 5-fold cross-validation to provide a fair and thorough comparison. The best-performing model is then fine-tuned to push its performance even further. Through this structured and transparent experimental design, we are able to draw meaningful conclusions about which algorithm is best suited for CAD prediction in a clinical setting.

\subsection{Large Language Models}
\begin{figure}[h]
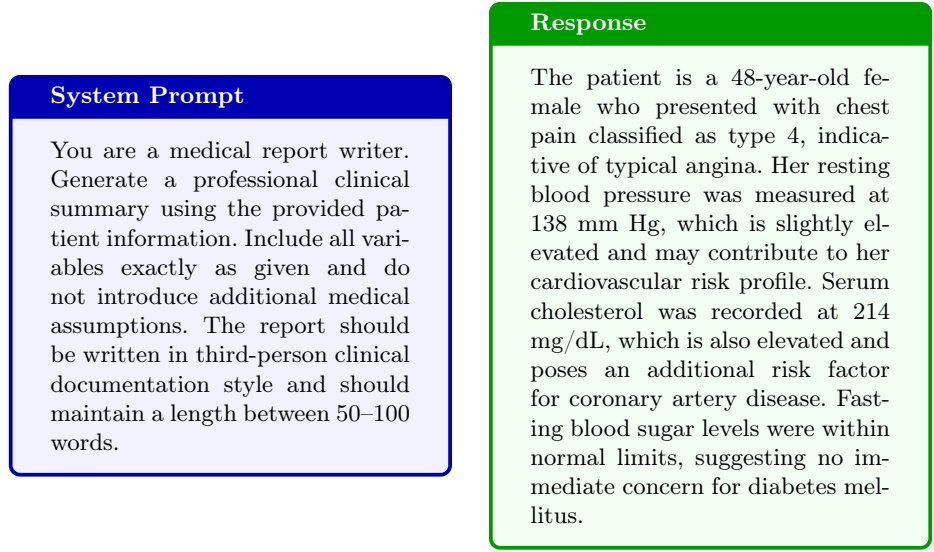

\centering

\begin{minipage}{0.48\textwidth}
\begin{tcolorbox}[
colback=blue!5,
colframe=blue!70!black,
title=System Prompt,
fonttitle=\bfseries
]

You are a medical report writer. Generate a professional clinical summary using the provided patient information. Include all variables exactly as given and do not introduce additional medical assumptions. The report should be written in third-person clinical documentation style and should maintain a length between 50--100 words.

\end{tcolorbox}
\end{minipage}
\hfill
\begin{minipage}{0.48\textwidth}
\begin{tcolorbox}[
colback=green!5,
colframe=green!60!black,
title=Response,
fonttitle=\bfseries
]
The patient is a 48-year-old female who presented with chest pain classified as type 4, indicative of typical angina. Her resting blood pressure was measured at 138 mm Hg, which is slightly elevated and may contribute to her cardiovascular risk profile. Serum cholesterol was recorded at 214 mg/dL, which is also elevated and poses an additional risk factor for coronary artery disease. Fasting blood sugar levels were within normal limits, suggesting no immediate concern for diabetes mellitus.

\end{tcolorbox}
\end{minipage}
\caption{Example of the prompt used for LLM-based clinical narrative generation and the corresponding generated response.}
\label{fig:fig10}
\end{figure}

The dataset consists of structured patient records with two columns, where patient data contains natural language descriptions of the clinical attributes shown in Fig. \ref{fig:fig10} and label represents the binary heart disease classification target. A balanced subset of 100 records is sampled equally from both classes to eliminate class imbalance bias, with labels normalized to a consistent format beforehand. We split the data into 70 training samples used as the few-shot example pool and 30 test samples kept fixed across all prompting conditions for a fair comparison. A set of shared helper functions is defined identically to ensure that prompt structure remains consistent across GPT and Gemini, so that performance differences reflect the models rather than input variations.

 In the zero-shot scenario, the model is presented with the patient's clinical data, solely on the pre-trained knowledge of the model, to generate a prediction. In the few-shot scenario, a set of patient examples is given to the patient's data, allowing the model to learn through examples. We draw an equal number of examples from each class and shuffle them to avoid ordering bias, while we evaluate precision, recall, F1-score, and accuracy, and plot a confusion matrix for each experiment. In both cases, temperature is set to 0 to enforce deterministic outputs. The Gemini additionally disables all safety filters to prevent content moderation from blocking medical classification responses. We define a retry wrapper function exclusively in Gemini to automatically re-attempt failed API calls up to two times, improving inference reliability. 
\section{Results and Discussion}
\subsection{Performance Analysis of Machine Learning Models}
\subsubsection{Model Comparison:}
\begin{figure}
    \centering
    \includegraphics[width=0.99\linewidth]{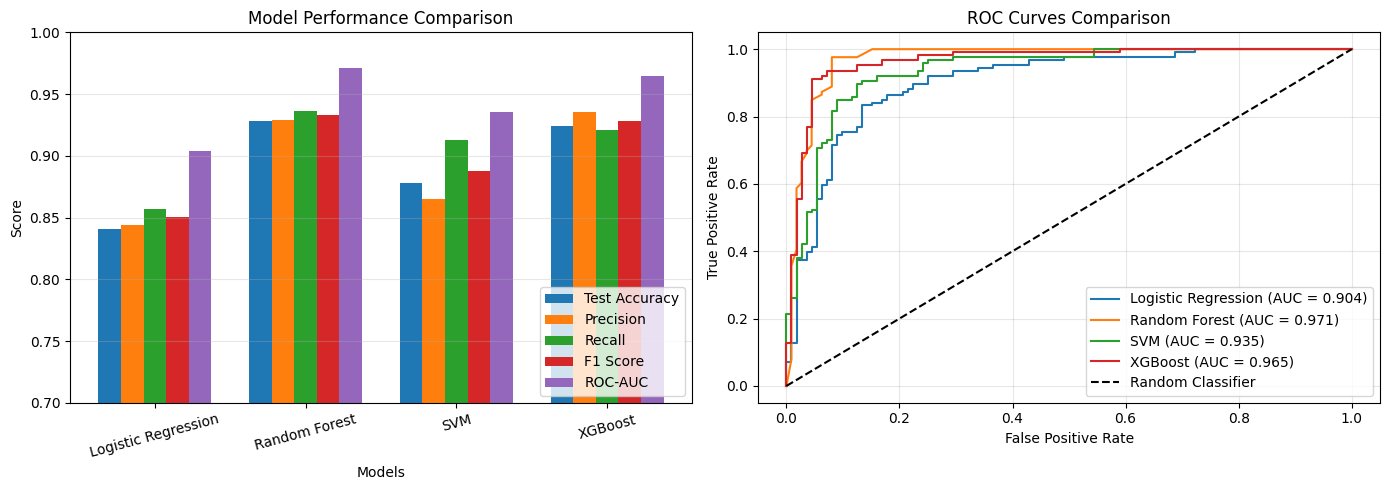}
    \caption{Comparison of Machine Learning Models}
    \label{fig:fig3}
\end{figure}

The four trained models are compared to find the best one. Graphs in Fig. \ref{fig:fig3} show test accuracy, precision, recall, and F1-score for each model. ROC curves for the models are shown together, with AUC values noted in the legend. Random Forest has the highest test accuracy (92.86\%) and ROC-AUC (97.07\%). XGBoost follows with 92.44\% accuracy and 96.49\% AUC. SVM and Logistic Regression show lower but still strong results. This confirms that ensemble methods outperform the linear baseline on this dataset.

\subsubsection{Feature Importance Analysis:}
\begin{figure}
    \centering
    \includegraphics[width=0.92\linewidth]{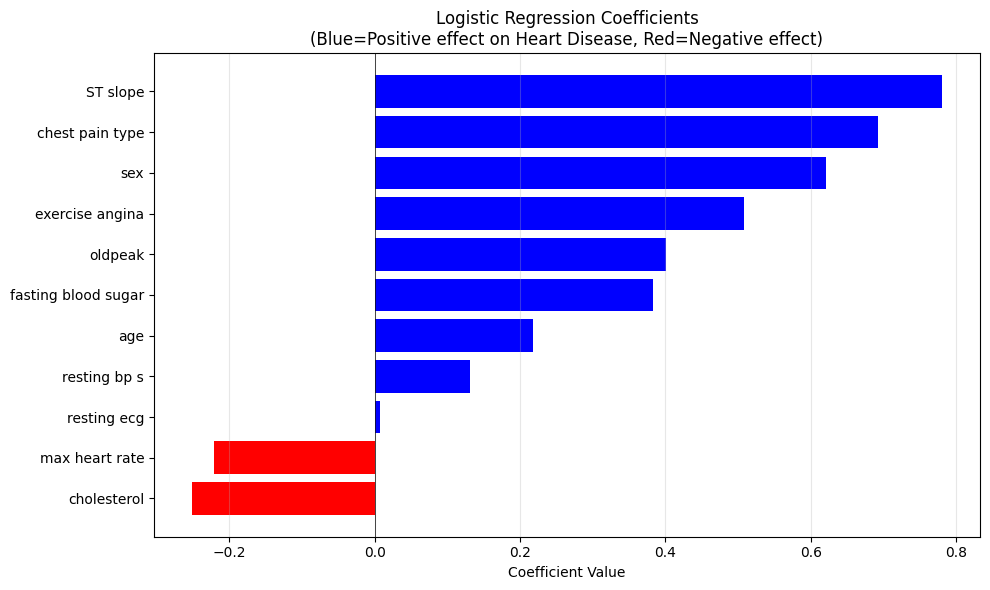}
    \caption{Logistic Regression coefficients}
    \label{fig:fig4}
\end{figure}

For determining the predictive nature of the trained models, we do the feature importance analysis. Random Forest and XGBoost models, being tree-based models, provide feature importance scores, which indicate the level of contribution of each feature to the final prediction. Additionally, the coefficients of the Logistic Regression model shown in Fig. \ref{fig:fig4} are used to understand the positive or negative relationship of clinical features with CAD risk. This enables the identification of the key cardiovascular risk factors influencing the model predictions.

\subsubsection{Hyperparameter Tuning:}
\begin{figure}
    \centering
    \includegraphics[width=0.7\linewidth]{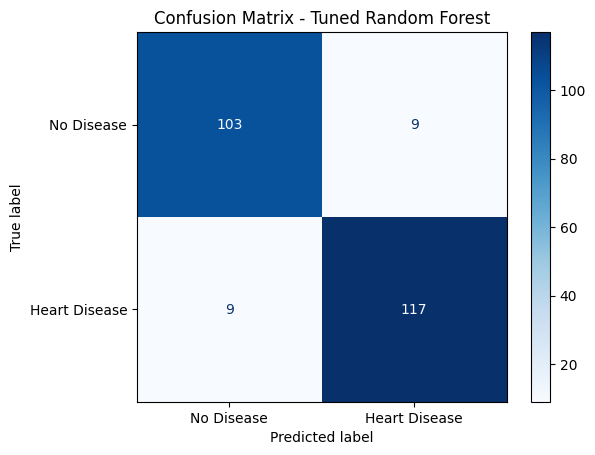}
    \caption{Confusion Matrix of tuned Random Forest}
    \label{fig:fig5}
\end{figure}

Random Forest achieves the highest test accuracy among all models, so it is selected for hyperparameter optimization. GridSearchCV is employed with 5-fold cross-validation to exhaustively search over a predefined grid of hyperparameter combinations. The parameter grid includes: number of estimators (100, 200, 300), maximum tree depth (None, 5, 10, 20), minimum samples required to split an internal node (2, 5, 10), minimum samples required at a leaf node (1, 2, 4), and whether bootstrap sampling is used (True, False). This results in 216 unique hyperparameter combinations, totaling 1,080 model fits. The tuned model achieves a test accuracy of 92.44\%, precision of 92.86\%, recall of 92.86\%, F1-score of 92.86\%, and ROC-AUC of 97.08\%. A confusion matrix is plotted for the tuned model to visualize prediction outcomes on the test set. Fig. \ref{fig:fig5} is the confusion matrix that visualizes the predicted outcomes.

\subsubsection{Prediction Function:}

We implement a prediction function, and this is used to show how the trained model may be applied to new patient data.
The prediction function is designed to take clinical attributes as input, apply the feature scaler, and make a prediction for the presence or absence of the disease. It will also provide a probability estimate for the prediction. 

\subsection{Performance Analysis of Large Language Models}
\subsubsection{Zero‑Shot Classification:}
We ask the model to classify patients without seeing any labeled examples first. Only the clinical story of test patients is entered into a prompt and sent to LLM. The prediction (heart disease or no heart disease) that the model provides is cleaned up into a standard label and stored. A short pause is inserted between requests so the API is not overloaded. 
After all the test predictions are collected, a helper function computes accuracy, precision, recall, and F1-score. It then draws a confusion matrix, so it is easy to see which cases the model gets right and where it makes mistakes. Same experiment is done for GPT and Gemini.
\begin{figure}
    \centering
    \includegraphics[width=0.98\linewidth]{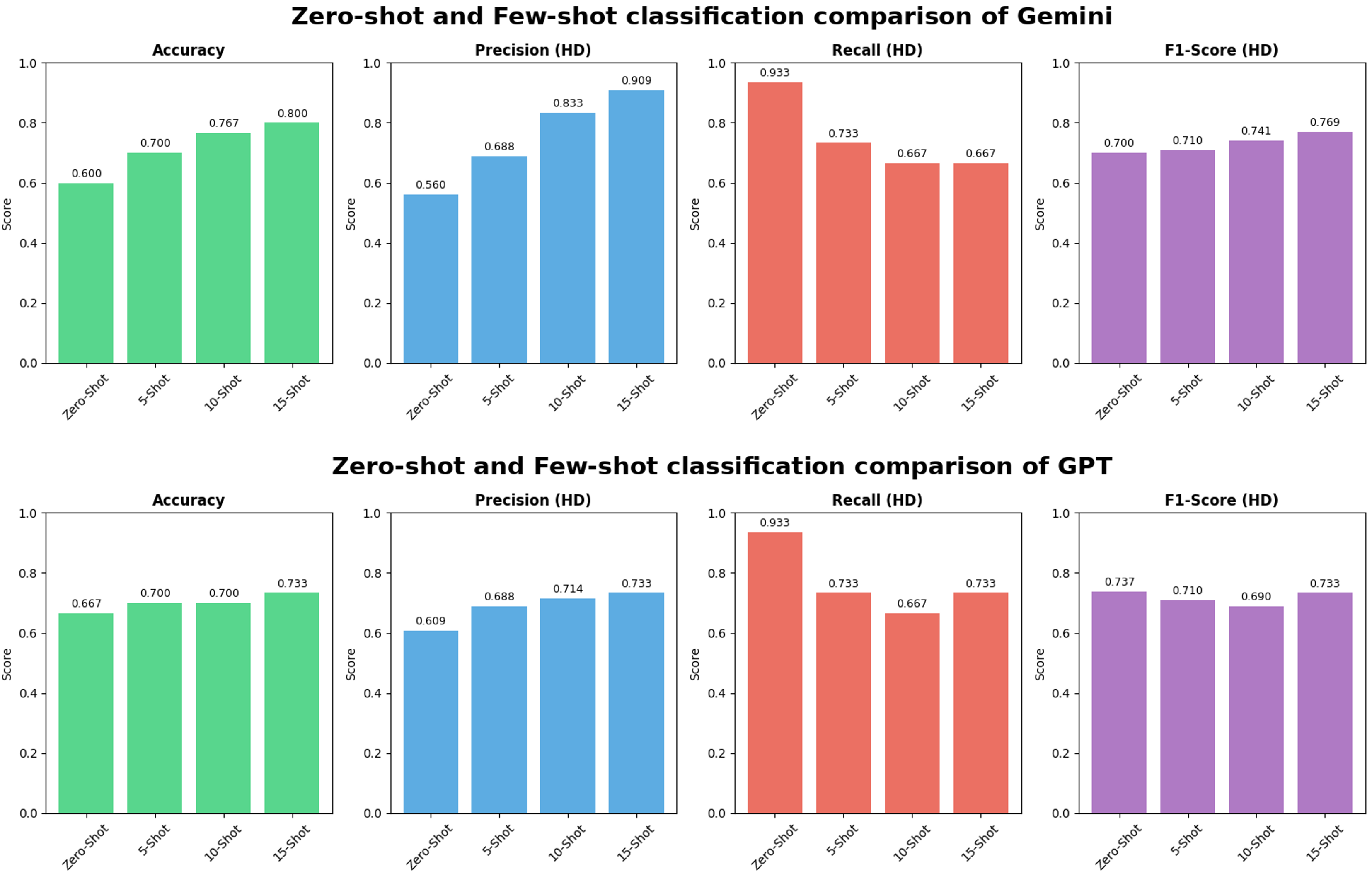}
    \caption{Model Comparison of LLMs}
    \label{fig:fig7}
\end{figure}
\subsubsection{Few-Shot Classification:}
Under a 5-shot prompting situation, we ask the model once again to do the same experiment. 5 balanced labeled examples (2 or 3 from each class) are selected from the training set and given to the prompt before each test patient. To provide a direct and equitable comparison, the same 30 test patients used in the zero-shot experiment are evaluated here. The same metrics and confusion matrix visualization are used to gather and assess predictions. We repeat the same experiment for 10 and 15 shot prompts by increasing the number of in-context examples. The same experiment is done for GPT and Gemini. The model gets more examples, which helps it to understand the classification boundary better.

\subsubsection{Model Comparison:}
To provide a comprehensive evaluation of the prompting tactics, we present the performance findings of all four conditions, Zero-Shot, 5-Shot, 10-Shot, and 15-Shot, in a unified summary for each model. For ensuring comparability between all conditions, four key classification metrics are reported for each prompting condition: accuracy, precision, recall, and F1-score Fig. \ref{fig:fig7}. These metrics are all calculated against the same fixed test set. These comparative analyses are carried out independently for both LLMs. It enables cross-model variations in the effectiveness of zero-shot and few-shot prompting to be examined alongside within-model trends. The performance comparison of machine learning models and large language models is summarized in Table~\ref{tab:cad_comparison}.






\begin{table}[h]
\centering
\caption{Performance comparison of ML models and LLM-based methods for CAD prediction}
\begin{tabular}{c|ccccc}
\hline
\textbf{Category} & \textbf{Model} & \textbf{Accuracy} & \textbf{Precision} & \textbf{Recall} & \textbf{F1-score} \\
\hline

Conventional ML & Logistic Regression & 84.10 & 84.30 & 85.70 & 85.00 \\

& Random Forest & 92.80 & 92.80 & 93.60 & 93.30 \\
& SVM & 87.80 & 86.50 & 91.30 & 88.80 \\
& XGBoost & 92.30 & 93.60 & 92.10 & 92.80 \\
\hline

Gemini (LLM) & Zero-shot & 60.00 & 56.00 & 93.30 & 70.00 \\
& 5-shot & 70.00 & 68.80 & 73.30 & 71.00 \\
& 10-shot & 76.70 & 83.30 & 66.70 & 74.10 \\
& 15-shot & 80.00 & 90.90 & 66.70 & 76.90 \\
\hline

GPT (LLM) & Zero-shot & 66.70 & 60.90 & 93.30 & 73.70 \\
& 5-shot & 70.00 & 68.80 & 73.30 & 71.00 \\
& 10-shot & 70.00 & 71.40 & 66.70 & 69.00 \\
& 15-shot & 73.30 & 73.30 & 73.30 & 73.30 \\
\hline

\end{tabular}
\label{tab:cad_comparison}
\end{table}

The best result was obtained with the Random Forest algorithm among all the machine learning models used in this experiment. However, despite this advantage, LLM-based classification remains a more preferable approach in real-world clinical settings. This is because LLMs operate directly on natural language patient descriptions, meaning that sensitive numerical patient data, such as exact lab values, blood pressure readings, and diagnostic codes, do not need to be explicitly exposed or shared. This makes LLM-based prediction inherently more privacy-friendly, as the model can reason from narrative text alone without requiring access to raw structured records.
\section{Conclusion and Future Work}
This study proposes a hybrid approach for CAD prediction, combining structured clinical data, large language models (LLMs), and machine learning. Numerical clinical data is transformed into synthetic narratives using LLMs, with a validation method ensuring over 94\% consistency. Additionally, the effectiveness of LLM models in CAD classification from natural language-based patient descriptions under zero-shot and few-shot prompt conditions is proven. 

The framework can be tested on larger and more diverse clinical data to enhance its generalizability. A potential extension of this study can be to fine-tune domain-specific language models, such as ClinicalBERT or biomedical LLMs, to enhance the quality of narrative generation and classification. Furthermore, the incorporation of other data modalities, such as ECG data, imaging data, or electronic health records, can benefit the prediction of CAD. Finally, the framework's potential for application in cardiovascular disease risk assessment by medical professionals might be useful by integrating it into a clinical decision support system.

%
%
\bibliographystyle{splncs04}

\begin{thebibliography}{10}
\providecommand{\url}[1]{\texttt{#1}}
\providecommand{\urlprefix}{URL }
\providecommand{\doi}[1]{https://doi.org/#1}

\bibitem{shao2020coronary}
Shao, C., Wang, J., Tian, J., Tang, Y.d.: Coronary artery disease: from
  mechanism to clinical practice. Coronary Artery Disease: Therapeutics and
  Drug Discovery pp. 1--36 (2020)

\bibitem{baghdadi2023advanced}
Baghdadi, N.A., Farghaly~Abdelaliem, S.M., Malki, A., Gad, I., Ewis, A., Atlam,
  E.: Advanced machine learning techniques for cardiovascular disease early
  detection and diagnosis. Journal of Big Data  \textbf{10}(1), ~144 (2023)

\bibitem{gururaj2025recent}
Gururaj, H., Flammini, F., Kumar, R., Prema, N.: Recent Trends in Healthcare
  Innovation: Proceedings of the Annual International Conference on Recent
  Trends in Healthcare Innovation (AICRTHI 2024), Mysuru, India, October
  24th--25th, 2024. CRC Press (2025)

\bibitem{mehdi2025llm}
Mehdi, S., Kouah, S., Saighi, A., Zertal, S.: Llm-cardio: A large language
  model-based assistant for cardiovascular health inquiry and diagnostic
  support using wearable data. Ing{\'e}nierie des Syst{\`e}mes d'Information
  \textbf{30}(12) (2025)

\bibitem{yang2025llm}
Yang, H., Shen, Z., Shao, J., Men, L., Han, X., Dong, J.: Llm-augmented symptom
  analysis for cardiovascular disease risk prediction: A clinical nlp approach.
  In: 2025 19th International Conference on Complex Medical Engineering (CME).
  pp. 176--181. IEEE (2025)

\bibitem{li2025ai}
Li, A., Wang, Y., Chen, H.: Ai driven cardiovascular risk prediction using nlp
  and large language models for personalized medicine in athletes. SLAS
  technology  \textbf{32},  100286 (2025)

\bibitem{chen2025large}
Chen, C., Li, L., Beetz, M., Banerjee, A., Gupta, R., Grau, V.: Large language
  model-informed ecg dual attention network for heart failure risk prediction.
  IEEE transactions on big data  \textbf{11}(3),  948--960 (2025)

\bibitem{wang2025boosting}
Wang, X., Wang, H., Fu, R.: Boosting cardiovascular disease prediction
  accuracy: A hybrid ai strategy integrating llm-generated risk scores. In:
  Proceedings of the 2025 International Conference on Health Informatization
  and Data Analytics. pp. 179--185 (2025)

\bibitem{pan2025integrating}
Pan, J., Lee, S., Cheligeer, C., Martin, E.A., Riazi, K., Quan, H., Li, N.:
  Integrating large language models with human expertise for disease detection
  in electronic health records. Computers in Biology and Medicine
  \textbf{191},  110161 (2025)

\bibitem{Mila2026ScreeningPaL}
Mila, S.A., Maliha, J., Kabir, M.R., Das, A., Ray, S.: Screeningpal: Llm-nlp
  enabled early autism detection method from caregiver’s free-text input. In:
  AMIA Amplify Informatics Conference (2026)

\bibitem{mondal2026ai}
Mondal, H.S., Feng, Y., Moushi, O.M., Birbilis, N.: Ai-driven biosensors for
  cardiac care: parametric-tuned modeling with extensive literature review.
  Iran Journal of Computer Science  \textbf{9}(1), ~11 (2026)

\bibitem{ahmed2025primer}
Ahmed, M., Lam, J., Chow, A., Chow, C.M.: A primer on large language models
  (llms) and chatgpt for cardiovascular healthcare professionals. CJC open
  \textbf{7}(5),  660--666 (2025)

\bibitem{dinh2019data}
Dinh, A., Miertschin, S., Young, A., Mohanty, S.D.: A data-driven approach to
  predicting diabetes and cardiovascular disease with machine learning. BMC
  medical informatics and decision making  \textbf{19}(1), ~211 (2019)

\end{thebibliography}

\end{document}